\documentclass{article}

\PassOptionsToPackage{dvipsnames,table}{xcolor}
\usepackage{iclr2027_conference,times}

\usepackage{amsmath,amsfonts,bm}

\def\eqref#1{equation~\ref{#1}}

\def\1{\bm{1}}

\DeclareMathAlphabet{\mathsfit}{\encodingdefault}{\sfdefault}{m}{sl}
\SetMathAlphabet{\mathsfit}{bold}{\encodingdefault}{\sfdefault}{bx}{n}

\usepackage{amsmath,amssymb,mathtools}
\usepackage{microtype}     
\usepackage{booktabs}
\usepackage{multirow}
\usepackage{makecell}      
\usepackage{graphicx}
\usepackage{float}
\usepackage{url}
\usepackage{enumitem}
\usepackage{algorithm}
\usepackage{algorithmic}
\usepackage{pifont}        
\usepackage{tcolorbox}     
\tcbuselibrary{breakable}

\definecolor{BrightTeal}{HTML}{45B8AC}
\definecolor{darkred}{RGB}{178,34,34}
\definecolor{questionbg}{RGB}{235,249,255}     
\definecolor{questionframe}{RGB}{135,190,230}  
\definecolor{insightbg}{RGB}{255,250,240}      
\definecolor{insightframe}{RGB}{200,160,100}   

\usepackage[
  colorlinks=true,
  linkcolor=BrightTeal,
  citecolor=Violet,
  urlcolor=Violet
]{hyperref}

\newcommand{\methodname}{\textsc{AscendEvolve}}

\newcommand{\pcref}[1]{\hyperref[#1]{\textcolor{BrightTeal}{\S\ref*{#1}}}}

\definecolor{obscolor}{HTML}{B22222}       
\definecolor{posgreen}{HTML}{2E7D32}       
\definecolor{tableheadbg}{HTML}{EAF1F8}    
\definecolor{ourowbg}{HTML}{F5F5F5}        
\newcommand{\obs}[1]{\textcolor{obscolor}{\textbf{#1}}}   
\newcommand{\dmark}[1]{\ding{#1}}                          
\newcommand{\posdelta}[1]{\textcolor{posgreen}{#1}}        

\newtcolorbox{questionbox}{
  colback=questionbg, colframe=questionframe,
  boxrule=0.4pt, arc=1pt,
  left=6pt, right=6pt, top=4pt, bottom=4pt,
  fonttitle=\bfseries, coltitle=black,
  breakable
}

\newtcolorbox{insightbox}{
  colback=insightbg, colframe=insightframe,
  boxrule=0.4pt, arc=1pt,
  left=6pt, right=6pt, top=4pt, bottom=4pt,
  breakable
}

\title{Beyond Experience: Knowledge-Guided Evolution for NPU Kernel Generation}

\author{%
\normalfont
Xi Chen$^{1,2,*}$ \quad Ming Shen$^{1,3,*}$ \quad Haotian Yan$^{1,4}$ \quad
Ruoyu Sun$^{5,1}$ \quad Tianjun Yao$^{1}$\\
\textnormal{$^{1}$Shenzhen Loop Area Institute (SLAI)}\\
\textnormal{$^{2}$Guangzhou University}\\
\textnormal{$^{3}$Shanghai Jiao Tong University}\\
\textnormal{$^{4}$Southwestern University of Finance and Economics}\\
\textnormal{$^{5}$The Chinese University of Hong Kong, Shenzhen}\\
\textnormal{*Equal contribution}%
}

\iclrfinalcopy

\begin{document}

\maketitle
\lhead{}

\begin{abstract}
LLM agents have become capable kernel engineers on mainstream GPU stacks, but
they struggle on emerging accelerators such as Ascend NPUs, whose programming
model is barely represented in pretraining data. Recent NPU agents address this
gap largely through experience: they accumulate memory from their own
generation and optimization trials and reuse it on later tasks. We argue that
experience alone is not enough for kernel generation. An agent can only
remember what it has tried, and it rarely tries a mechanism it does not know.
As a result, such memory offers little help when a new operator calls for a
different mechanism, and even familiar operators raise unseen situations as
optimization proceeds. We present \methodname{}, which grounds every step of an
evolutionary kernel search in a large external knowledge base built from
platform documentation, toolkit headers, and operator repositories. Because
this corpus far exceeds any context window, \methodname{} indexes it and routes
each agent to a small, task-specific slice of evidence instead of the corpus
itself. The retrieved knowledge seeds diverse implementation strategies, guides
repair when a candidate fails, and constrains how strong candidates are
recombined. On 53 Ascend~C operators from CANN-Bench spanning four difficulty
levels, \methodname{} achieves an overall score of 71.39, improving over
Reflexion and OpenEvolve by 11.6\% and 17.2\%.
\end{abstract}

\section{Introduction}
\label{sec:intro}

Trial-and-error agents learn from the programs they actually generate and evaluate. As a result, a memory built from past trials is naturally limited by what the agent has explored: it can reuse and refine discovered mechanisms, but does not guarantee access to mechanisms that have never appeared in its search. This motivates knowledge beyond self-generated experience.

Recent work has demonstrated substantial progress in LLM-based kernel
generation and optimization for GPUs. On CUDA and related programming stacks,
iterative workflows that combine generation with compilation, execution, and
profiling feedback can progressively improve both correctness and performance
\citep{ouyang2025kernelbench,zhang2025cudaforge,wei2025astra,baronio2025kevin}.
These results establish iterative execution feedback as an effective ingredient
for kernel optimization on mature GPU ecosystems. The problem becomes
considerably more challenging on specialized accelerator programming stacks for
which general-purpose LLMs have demonstrated substantially lower capability---a
difficulty well documented for domain-specific programming languages more
broadly \citep{joel2025surveyllmbasedcodegeneration}. On Huawei's Ascend NPU
\citep{9407221}, recent work reports that general-purpose LLMs achieve
very low success rates on complex kernel-generation tasks without
domain-specific adaptation \citep{cao2026ascendkernelgen}.

\begin{figure}[t]
  \centering
  \includegraphics[width=\textwidth]{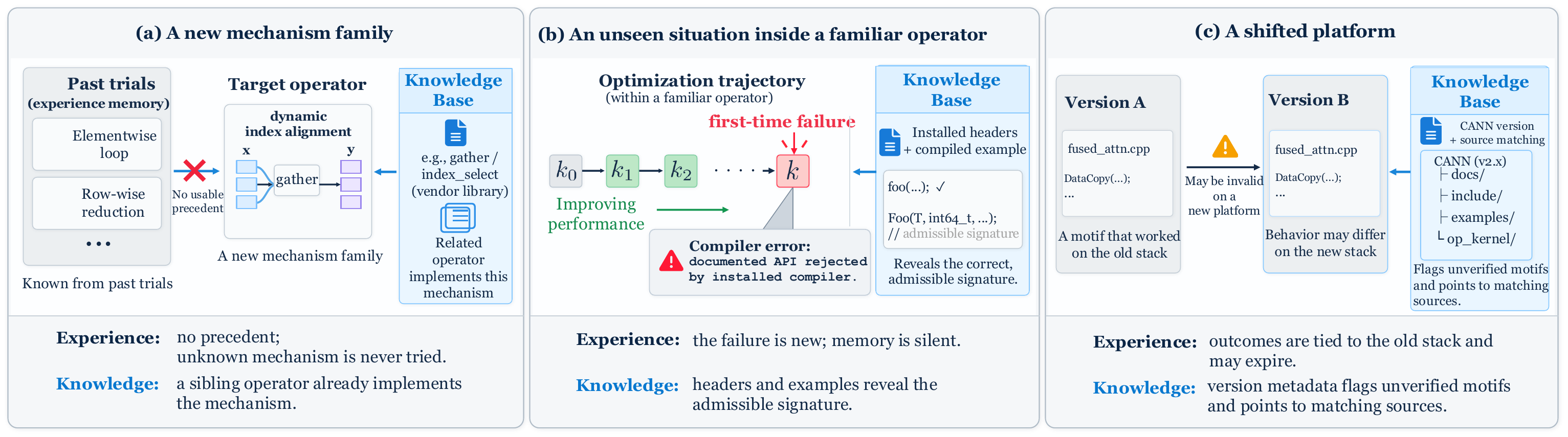}
  \caption{Three situations in which an agent's own trials offer no usable
  precedent while primary platform sources do: (a)~a mechanism family absent
  from past trials, (b)~a first-time failure midway through optimization, and
  (c)~a motif carried across toolkit versions. Illustrative, not measured.}
  \label{fig:motivation}
\end{figure}

Recent systems have addressed this difficulty in part by accumulating reusable
experience from previous generation and optimization trajectories. Value-driven
memory substantially improves functional correctness over zero-shot prompting
\citep{zheng2026towards}, and experience mined from optimization trajectories
yields consistent speedups on existing operators
\citep{wu2026ascendoptimizer,zhang2026accelopt}. These results provide strong
evidence that reusable experience can substantially improve both feasibility
and optimization efficiency. However, these systems necessarily construct their
reusable evidence from trajectories that the agent has already proposed and
evaluated. This suggests a complementary limitation: the usefulness of
experience depends not only on how much is retained, but also on what the agent
has managed to explore.

\begin{insightbox}
We refer to this coverage constraint as
\emph{experience closure}: the reusable evidence in a self-generated memory is
derived from trajectories the agent has actually proposed and evaluated. As a
result, such memory may provide limited guidance for mechanisms that have not
yet appeared in those trajectories---particularly on platforms where the
model's familiarity with the programming stack is limited to begin with.
\end{insightbox}

Figure~\ref{fig:motivation} shows three settings where self-generated
experience provides limited guidance:
\begin{enumerate}[label=(\roman*),leftmargin=*,itemsep=2pt,topsep=4pt]
\item \textbf{Coverage gap.} An operator may
require a mechanism family that no previous trajectory has explored; since such
a mechanism is rarely proposed, it is also unlikely to be recorded.
\item \textbf{Repair gap.} A familiar operator may encounter, midway through optimization,
a constraint that has not appeared before---compiler diagnostics often identify
the violated constraint without directly providing a valid alternative
implementation.
\item \textbf{Validity gap.} Experience collected under one
software or hardware configuration may not transfer unchanged to another.
\end{enumerate} Self-generated experience is not the only source of
implementation evidence. Accelerator ecosystems expose additional information
through programming guides, API definitions, installed headers, reference
implementations, and vendor-maintained operator libraries~\citep{cutlass,zhao2021akg}. These sources can
provide mechanism and constraint information before the agent has encountered
the corresponding trajectory itself. We therefore view primary platform
knowledge and execution experience as complementary: platform knowledge can
broaden the set of mechanisms the agent considers, while execution experience
provides empirical evidence about which candidates are correct and efficient on
the target device \citep{wen2026hawk}. Our goal is not to replace trial-and-error with retrieval, but to inform it.

We present \methodname{}, a system that grounds every stage of an evolutionary
kernel search in primary platform sources. The design follows three principles:
\begin{itemize}[leftmargin=*,itemsep=2pt,topsep=4pt,label={}]
\item \textcolor{obscolor}{\dmark{172}}~\textbf{Hierarchical and version-aware.} Platform information is organized hierarchically and with version
awareness: a shared, versioned corpus of documentation, installed headers, and
vendor operator libraries is distilled into per-operator evidence before search
begins, avoiding repeated processing of a large heterogeneous corpus.
\item \textcolor{obscolor}{\dmark{173}}~\textbf{Demand-driven retrieval.} Rather than placing the corpus in context,
\methodname{} routes each concrete implementation or optimization question to a
small set of relevant sources.
\item \textcolor{obscolor}{\dmark{174}}~\textbf{Bidirectional interaction.} Retrieved platform evidence and execution
experience interact bidirectionally: retrieved mechanisms seed candidate
implementations, each trial records an empirical verdict on the knowledge it
was grounded in, and outcomes influence subsequent retrieval.
\end{itemize}
Because retrieved
evidence may suggest multiple plausible implementation mechanisms rather than a
single canonical solution, \methodname{} maintains several mechanism-distinct
implementation lines and progressively selects and recombines promising design
choices through evolutionary search.

On 53 Ascend~C operators spanning four difficulty levels,
\methodname{} achieves an overall score of \textbf{71.39}, improving over
Reflexion~\citep{shinn2023reflexion} and OpenEvolve~\citep{openevolve} by
\textbf{11.6\%} and \textbf{17.2\%}, respectively, with the
largest gains on the operator categories where prior agents have the
lowest correctness coverage. On L4 operators, \methodname{} reaches \textbf{35.0\%}
correctness coverage compared with 4.38\% for Reflexion and 1.25\% for
OpenEvolve. Removing the knowledge base nearly halves the L3 score,
showing how platform knowledge helps generate correct kernels.
Our contributions are as follows:

\begin{itemize}[leftmargin=*,itemsep=3pt,topsep=4pt,label={}]
\item \textcolor{obscolor}{\dmark{172}} \textbf{Experience closure.}
  We identify experience closure---a coverage limitation of
  experience-driven kernel agents on platforms outside the model's
  familiarity---and characterize three settings where it is most pronounced.
\item \textcolor{obscolor}{\dmark{173}} \textbf{Method.}
  We propose \methodname{}, which couples primary platform knowledge with
  procedural experience through hierarchical, demand-driven retrieval and
  bidirectional feedback within an evolutionary search that maintains
  mechanism-diverse implementation lines.
\item \textcolor{obscolor}{\dmark{174}} \textbf{Evaluation.}
  We evaluate on 53 Ascend~C operators from CANN-Bench~\citep{gao2026cann} and show
  that knowledge grounding yields substantial gains, particularly on
  hard operators where experience-only methods have very low correctness
  coverage. We isolate the contribution of the knowledge base through
  ablation.
\end{itemize}

\section{Problem Formulation}
\label{sec:problem}

\paragraph{Notation.}
Let $o$ denote an operator specified by an executable reference implementation and a set of test cases $\mathcal{T} = \{(\mathbf{x}_i, \mathbf{y}_i)\}_{i=1}^{N}$, where each $(\mathbf{x}_i, \mathbf{y}_i)$ is an input--output pair.
A candidate kernel $x$ is a compilable Ascend C program that implements $o$.
We write $C(x) \in [0,1]$ for the fraction of test cases that $x$ passes under a fixed numerical-tolerance criterion, and $L(x)$ for the wall-clock latency of $x$ measured on the target device under a fixed profiling protocol.
$\mathcal{K}_G$ is a shared, static knowledge base of platform documentation, vendor libraries and machine-observed facts.
$\mathcal{K}_o$ is a per-operator knowledge base that fixes the task contract and links it to $\mathcal{K}_G$.
$\mathcal{E}$ is a dynamic experience store that accumulates trial records during search.
$R(q \mid \mathcal{K}_o, \mathcal{K}_G, \mathcal{E})$ is a retrieval function that, given a concrete question $q$, returns a small \emph{evidence pack} $\mathcal{C}$ from the knowledge bases, filtered by experience.

\paragraph{Problem setup.}
The goal is to produce a kernel $x^\star$ that is correct on all test cases and as fast as possible on the target device. We compare candidates with a correctness-first fitness
\begin{equation}
F(x) = \bigl(C(x),\, -L(x)\bigr),
\label{eq:fitness}
\end{equation}
ordered lexicographically: latency is compared only between candidates of equal correctness.
The search is conducted by an LLM-based agent that, at each step $t$, conditions on an evidence pack
\begin{equation}
\mathcal{C}_t = R\bigl(q_t \mid \mathcal{K}_o, \mathcal{K}_G, \mathcal{E}_{<t}\bigr)
\label{eq:evidence}
\end{equation}
to propose, evaluate and refine candidates. The agent must maximize $F$ within $T$ LLM calls.

\section{Method}
\label{sec:method}

\subsection{Overview}
\label{sec:overview}

\begin{figure}[t]
  \centering
  \includegraphics[width=\textwidth]{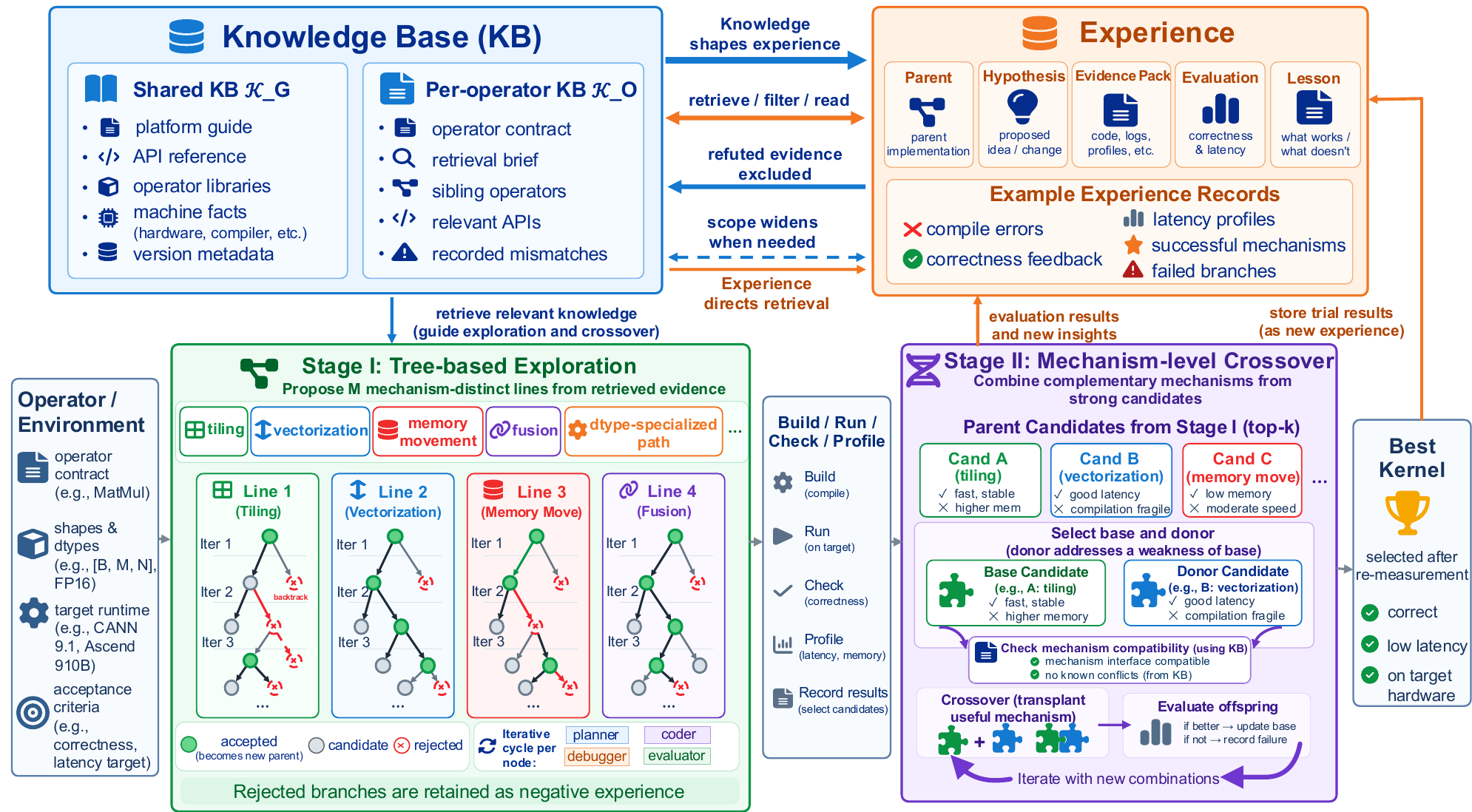}
  \caption{Overview of \methodname{}. Retrieval draws a task-specific evidence
  pack from $\mathcal{K}_G$ and $\mathcal{K}_o$. Stage~I explores $M$
  mechanism-distinct lines as trees, keeping rejected branches as negative
  experience; Stage~II transplants a donor mechanism into the best base after a
  compatibility check. Each trial writes a record that steers later retrieval.}
  \label{fig:framework}
\end{figure}

Existing agents search the space of candidate kernels with an LLM policy conditioned on experience $\mathcal{E}_{<t}$ gathered from earlier trials. On a platform the LLM barely knows, this conditioning is not enough: the policy can only revisit and recombine what it has already tried, and a mechanism that lies outside its prior is never sampled in the first place. \methodname{} adds a second source of information that does not depend on the agent's own behavior: every decision is conditioned on the evidence pack $\mathcal{C}_t$ (Eq.~\ref{eq:evidence}), which retrieves version-checked platform facts and expert implementation patterns from the knowledge bases $\mathcal{K}_G$ and $\mathcal{K}_o$, filtered by accumulated experience. The rest of this section describes how the knowledge bases are built (\S\ref{sec:kb}), how $R$ retrieves from a corpus far larger than any context window (\S\ref{sec:retrieval}), how knowledge and experience interact (\S\ref{sec:interaction}), and how both drive a two-stage evolutionary search (\S\ref{sec:search}). Figure~\ref{fig:framework} gives an overview.

\subsection{Knowledge Base Construction}
\label{sec:kb}

Writing a kernel requires answers to three questions: what the operator must compute, what the platform permits, and how experts implement similar operators on it. The first is specific to each task. The other two are shared across tasks. We organize knowledge along this line.

\paragraph{Shared knowledge base.}
$\mathcal{K}_G$ holds platform knowledge and is built from primary sources, not from hand-written summaries. It contains the official programming guide, API reference and compilable examples; the vendor's open-source operator libraries, which expose production implementations across operator families; and facts recorded from the target machine, such as installed headers and the compiler in use. Each source $d \in \mathcal{K}_G$ carries metadata $m(d)$ that records its toolkit version, hardware scope and trust level; this metadata induces an authority order $\succ$ used during retrieval (\S\ref{sec:retrieval}). $\mathcal{K}_G$ is immutable during search. Appendix~\ref{app:kb} lists the sources and their statistics.

\paragraph{Per-operator knowledge base.}
$\mathcal{K}_o$ fixes the task and tells the agent where to look. It has two parts. The \emph{contract} freezes the operator semantics, the executable reference, the shape and dtype inventory, the target environment and the acceptance criteria; it is the sole authority on what the kernel must compute. The \emph{retrieval brief} links the operator to $\mathcal{K}_G$: it names the operator family, the closest sibling implementations and the relevant API entries, and it records any disagreement found between these sources and the contract. $\mathcal{K}_o$ is produced fully automatically by a preparation pass that runs once per operator before search begins, with no human editing, and it stores references into $\mathcal{K}_G$, never copies. In one of our tasks, for example, the pass found that a closely related vendor operator places a numerical-stability constant differently from the executable reference. The brief records this mismatch, so later agents borrow the vendor's tiling without its arithmetic.
\subsection{Retrieval from a Large Knowledge Base}
\label{sec:retrieval}

$\mathcal{K}_G$ exceeds the context window by orders of magnitude, and most of it is irrelevant to any single decision. Two properties of kernel code make standard dense retrieval~\citep{lewis2021retrieval} a poor fit. Queries are frequently exact symbols, and a nearly identical passage from a mismatched toolkit version is actively harmful, since it leads to code that looks right and does not compile. We therefore design $R$ as a hierarchical, demand-driven procedure with three steps.

\textcolor{obscolor}{\dmark{172}}~\textbf{Route.} Retrieval is triggered by one concrete question $q_t$, such as how a given API handles tail elements, and never by the task as a whole. The retrieval brief in $\mathcal{K}_o$, backed by a global index over $\mathcal{K}_G$, maps $q_t$ to a short list of candidate sources. This step reduces the search space from the whole corpus to a handful of files.

\textcolor{obscolor}{\dmark{173}}~\textbf{Filter.} Candidates are checked against the target environment using $m(d)$. Sources of a mismatched version are either dropped or kept as explicitly labeled leads. When two candidates disagree, the authority order $\succ$ breaks the tie: facts observed on the target machine outrank official sources of the matching version, which outrank other versions and informal material. The disagreement itself is retained in the pack so that the agent is aware of it.

\textcolor{obscolor}{\dmark{174}}~\textbf{Read.} The agent reads the surviving sources directly, locating the relevant region by exact symbol or path. Implementation patterns are returned with their surrounding code, because a tiling loop is meaningless without the declarations it relies on.

If a question remains unresolved after these steps, the scope widens one level along the index. Each item in $\mathcal{C}_t$ keeps its source identity, so that decisions in the final kernel are traceable to evidence. Appendix~\ref{app:retrieval} gives the full procedure and an example pack.

\subsection{Interaction between Knowledge and Experience}
\label{sec:interaction}

Knowledge and experience answer different questions. Knowledge states what is possible and legal; experience states what worked, on this operator and this device. Neither is sufficient alone. Knowledge without experience cannot say which of several legal mechanisms is fastest, and experience without knowledge is confined to the mechanisms the LLM already knows. \methodname{} couples the two in both directions.

\paragraph{Knowledge shapes experience.}
Every trial starts from a hypothesis $h_t$ that is grounded in an evidence pack, and the record written after evaluation keeps this link:
\begin{equation}
e_t = \bigl(p_t,\; h_t,\; \mathcal{C}_t,\; F(x_t),\; \ell_t\bigr),
\label{eq:experience}
\end{equation}
where $p_t$ is the parent candidate and $\ell_t$ is a short lesson. An experience record is thus more than a score attached to a code change. It is an empirical verdict on a piece of knowledge in a specific context, such as ``the vectorized path from a sibling operator is valid here and reduces latency on large shapes'' or ``this API is documented but rejected by the installed compiler.'' Because trials are seeded by retrieved mechanisms, the experience the agent accumulates covers regions of the design space that it would not have reached from its prior.

\paragraph{Experience directs retrieval.}
The retrieval function in Eq.~\ref{eq:evidence} is conditioned on $\mathcal{E}_{<t}$: the outcome of the last trial determines the next question (a compilation error routes to headers, a numerical mismatch to the reference and dtype constraints, a regression to profiles); evidence refuted for this operator is excluded from later packs; and when accumulated failures show that the current evidence is insufficient, retrieval widens its scope. In this way experience handles situations that arise only during optimization and that no static brief could anticipate. When knowledge and measurement disagree, the on-device result overrides documentation for the task at hand; the verdict is written to the task-local experience, never to $\mathcal{K}_G$, keeping platform facts free of candidate-specific outcomes.

\subsection{Knowledge-Grounded Evolutionary Search}
\label{sec:search}

The search has two stages (Algorithm~\ref{alg:method}). Stage~I explores mechanism-distinct implementation lines, and Stage~II recombines the best of them.

\paragraph{Stage I: tree exploration.}
The planner first proposes $M$ implementation lines that differ in mechanism, for example in tiling scheme, vectorization strategy or dtype-specialized path. Diversity comes from the different mechanisms found in the retrieved evidence, not from prompt perturbation. Each line then runs for $K$ iterations. In one iteration, following the role decomposition common to iterative code agents~\citep{madaan2023self,shinn2023reflexion}, a planner selects an accepted parent and states one falsifiable hypothesis with its expected effect; a coder realizes it as a child candidate; a debugger applies bounded repairs that preserve the hypothesis; and an evaluator runs the build, correctness and latency gates in order. If the child improves $F$, it becomes the next parent. Otherwise the line backtracks to an earlier accepted node and the planner forms a different hypothesis. The history of a line is therefore a tree. Rejected branches are kept as negative experience~\citep{andrychowicz2017hindsight}, which prevents a harmful change from contaminating later candidates and records which assumptions are already falsified.

\paragraph{Stage II: mechanism-level crossover.}
Stage~I returns the best correct candidate of every line together with a summary of its mechanism, its per-case latency and its known weaknesses. Stage~II treats these candidates as a population. The best candidate under $F$ serves as the base, and a donor is selected only if its mechanism addresses a specific weakness of the base, such as a slow shape regime or an inefficient tail path. Here the two information sources play complementary roles again: experience identifies \emph{which} mechanism wins in which regime, and the knowledge base verifies \emph{whether} the two mechanisms can coexist, in terms of tiling structure, dtype behavior and platform constraints. Crossover transplants the donor mechanism into the base instead of concatenating implementations. An offspring that improves $F$ becomes the new base, and a failed combination is recorded so that the pairing is not tried again. The search ends when the budget $B$ is spent or no compatible hypothesis remains, and the finalists are re-measured under the same protocol before the best kernel is returned.

\begin{algorithm}[t]
\caption{\methodname{}}
\label{alg:method}
\begin{algorithmic}[1]
\REQUIRE operator $o$, shared KB $\mathcal{K}_G$, lines $M$, iterations $K$, crossover budget $B$
\STATE $\mathcal{K}_o \leftarrow \textsc{Prepare}(o, \mathcal{K}_G)$; \quad $\mathcal{E} \leftarrow \emptyset$
\STATE propose $M$ mechanism-distinct lines from $R(\cdot \mid \mathcal{K}_o, \mathcal{K}_G, \mathcal{E})$
\STATE \COMMENT{Stage~I: tree exploration}
\FOR{each line $i = 1,\dots,M$ \textbf{and} $k = 1,\dots,K$}
    \STATE select an accepted parent $p$; form question $q$ from $\mathcal{E}$
    \STATE $\mathcal{C} \leftarrow R(q \mid \mathcal{K}_o, \mathcal{K}_G, \mathcal{E})$ \hfill $\triangleright$ route, filter, read
    \STATE plan hypothesis $h$; generate child $x$; repair if needed
    \STATE evaluate $F(x)$; \quad $\mathcal{E} \leftarrow \mathcal{E} \cup \{(p, h, \mathcal{C}, F(x), \ell)\}$
    \STATE accept $x$ as new parent if $F$ improves, else backtrack
\ENDFOR
\STATE $x^\star \leftarrow$ best candidate over all lines
\STATE \COMMENT{Stage~II: mechanism-level crossover}
\FOR{$b = 1,\dots,B$}
    \STATE choose donor $x'$ whose mechanism addresses a weakness of $x^\star$ \hfill $\triangleright$ from $\mathcal{E}$
    \STATE \textbf{if} mechanisms are incompatible under $\mathcal{K}_o, \mathcal{K}_G$ \textbf{then continue}
    \STATE $x'' \leftarrow \textsc{Crossover}(x^\star, x')$; evaluate; update $\mathcal{E}$
    \STATE \textbf{if} $F(x'') \succ F(x^\star)$ \textbf{then} $x^\star \leftarrow x''$
\ENDFOR
\RETURN $x^\star$ after re-measurement
\end{algorithmic}
\end{algorithm}

\section{Experiments}
\label{sec:experiments}

\subsection{Experimental Setup}
\label{sec:exp-setup}

\paragraph{Dataset.}
We evaluate on CANN-Bench~\citep{gao2026cann}, using all 53 public operators across four difficulty levels, with 20 test cases per operator (1{,}060 cases in total). The benchmark progresses from relatively simple elementwise and Vector-Core operators (L1--L2) to increasingly complex fused Vector--Cube workloads (L3--L4), allowing us to examine how the methods behave as implementation requirements become more specialized.

\paragraph{Baselines.}
We compare against two test-time search paradigms that form a natural progression. \textbf{Reflexion}~\citep{shinn2023reflexion} represents sequential experience-based refinement using execution feedback. \textbf{OpenEvolve}~\citep{openevolve} represents population-based evolutionary search. All methods use GPT-5.5 with medium reasoning effort as the backbone, with Codex as Reflexion's agent harness, and the same benchmark inputs and evaluator throughout.

\paragraph{Evaluation.}
We follow the staged evaluation logic of CANN-Bench~\citep{gao2026cann} but recalibrate its performance component for our target NPU. \emph{Build} is a binary result for each operator's selected kernel. \emph{Runtime} measures whether each test case executes without failure, and \emph{correctness} checks its output against the reference implementation. A case receives \emph{performance} credit only if it passes correctness and both the candidate and \texttt{torch\_npu} have valid positive latency measurements. Build coverage is operator-level; all other coverage is case-level.

Since the published CANN-Bench performance anchors are hardware-specific, we use the latency of the corresponding \texttt{torch\_npu} implementation as a common baseline. Let $b_o$ indicate whether operator $o$ builds, and let $r_{o,i}$ and $c_{o,i}$ indicate whether case $i$ runs and passes correctness, respectively. We define $R_{o,i}=b_o r_{o,i}$ and $A_{o,i}=R_{o,i}c_{o,i}$, so a build failure gives all 20 cases zero credit. The first score component is $\sum_{i=1}^{20}R_{o,i}$: each case that runs after a successful build earns one point. The score for operator $o$ is
\begin{equation}
S_o
=
100\left[
\frac{1}{20}
\sum_{i=1}^{20}
\left(
0.2R_{o,i}+A_{o,i}\left(0.3+0.5P_{o,i}\right)
\right)
\right],
\label{eq:score}
\end{equation}
where $P_{o,i}=\operatorname{clip}\!\bigl(\sqrt{T_{\mathrm{torch},o,i}/T_{\mathrm{cand},o,i}},\;0.5,\;4\bigr)$ when both latencies are valid and positive, and $P_{o,i}=0$ otherwise. Thus, 100 corresponds to full correctness with \texttt{torch\_npu}-level performance when all cases have valid latencies, while faster correct kernels may score above 100. We macro-average operator scores so each contributes equally.

\subsection{Main Results}
\label{sec:main-results}

\begin{table}[t]
  \caption{Main results on CANN-Bench. Build is reported over operators; runtime, correctness and performance coverage are reported over test cases. Best score per level in \textbf{bold}.}
  \label{tab:main-results}
  \centering
  \small
  \scalebox{0.95}{%
  \begin{tabular}{llrrrrr}
    \toprule
    \rowcolor{tableheadbg}
    Level & Method
          & \thead[r]{Build\\(\%)}
          & \thead[r]{Runtime\\cases (\%)}
          & \thead[r]{Correct\\cases (\%)}
          & \thead[r]{Perf.\\cases (\%)}
          & \thead[r]{Score\\$\uparrow$} \\
    \midrule
    \multirow{3}{*}{L1 (8)}
      & Reflexion     & 100.00 & 100.00 & 96.25  & 96.25  & 109.58 \\
      & OpenEvolve    & 100.00 & 100.00 & 100.00 & 100.00 & 100.37 \\
      \rowcolor{ourowbg}
      & \methodname{} & 100.00 & 100.00 & 95.63  & 95.63  & \textbf{111.18} \\
    \midrule
    \multirow{3}{*}{L2 (16)}
      & Reflexion     & 100.00 & 100.00 & 99.06  & 91.56  & 81.38 \\
      & OpenEvolve    & 100.00 & 100.00 & 77.50  & 65.94  & 71.27 \\
      \rowcolor{ourowbg}
      & \methodname{} & 100.00 & 100.00 & 85.63  & 78.75  & \textbf{87.84} \\
    \midrule
    \multirow{3}{*}{L3 (21)}
      & Reflexion     & 100.00 & 85.48  & 61.19  & 41.90  & 53.56 \\
      & OpenEvolve    & 85.71  & 79.76  & 66.43  & 47.86  & \textbf{57.84} \\
      \rowcolor{ourowbg}
      & \methodname{} & 100.00 & 90.48  & 59.76  & 48.33  & 56.94 \\
    \midrule
    \multirow{3}{*}{L4 (8)}
      & Reflexion     & 87.50  & 48.13  & 4.38   & 0.00   & 10.94 \\
      & OpenEvolve    & 50.00  & 41.25  & 1.25   & 0.63   & 8.78 \\
      \rowcolor{ourowbg}
      & \methodname{} & 87.50  & 86.88  & 35.00  & 35.00  & \textbf{36.63} \\
    \midrule
    \multirow{3}{*}{Overall (53)}
      & Reflexion     & 98.11  & 86.42  & 69.34  & 58.77  & 63.98 \\
      & OpenEvolve    & 86.79  & 83.11  & 65.00  & 54.06  & 60.91 \\
      \rowcolor{ourowbg}
      & \methodname{} & 98.11  & 94.25  & 69.25  & 62.64  & \textbf{71.39} \\
    \bottomrule
  \end{tabular}
  }
\end{table}

\methodname{} achieves the highest overall score (71.39), improving over Reflexion and OpenEvolve by 11.6\% and 17.2\%, respectively (Table~\ref{tab:main-results}). The source of the gain changes with difficulty. On L1--L2, where executability is already high, improvements come from stronger optimization rather than broader coverage. On L4, the bottleneck shifts to implementation feasibility: \methodname{} reaches 35.0\% correctness coverage, compared with 4.38\% for Reflexion and 1.25\% for OpenEvolve. L3 is more mixed: OpenEvolve obtains a slightly higher final score, while \methodname{} achieves substantially higher build and runtime coverage.

\obs{Observation~1.} External knowledge becomes increasingly important when the challenge is no longer merely optimizing a valid implementation, but \textbf{finding a valid implementation mechanism}.

\subsection{Where Do the Gains Come From?}
\label{sec:score-decomposition}

\begin{table}[t]
  \caption{Overall score decomposition. Successful runs contribute $\sum_i R_{o,i}$ points per operator, one per case that runs after a successful build.}
  \label{tab:overall-score-decomposition}
  \centering
  \small
  \begin{tabular}{lrrrrrrrr}
    \toprule
    \rowcolor{tableheadbg}
    Method
      & \thead[r]{Successful\\runs (pts)}
      & \thead[r]{Correct.\\score}
      & \thead[r]{Perf.\\score}
      & \thead[r]{Score\\$\uparrow$}
      & \thead[r]{Build\\(\%)}
      & \thead[r]{Runtime\\cases (\%)}
      & \thead[r]{Correct\\cases (\%)}
      & \thead[r]{Perf.\\cases (\%)} \\
    \midrule
    \rowcolor{ourowbg}
    \methodname{}
      & \textbf{18.85} & 20.77 & \textbf{31.77} & \textbf{71.39}
      & 98.11 & \textbf{94.25} & 69.25 & \textbf{62.64} \\
    Reflexion
      & 17.28 & \textbf{20.80} & 25.90 & 63.98
      & 98.11 & 86.42 & \textbf{69.34} & 58.77 \\
    OpenEvolve
      & 16.62 & 19.50 & 24.79 & 60.91
      & 86.79 & 83.11 & 65.00 & 54.06 \\
    \bottomrule
  \end{tabular}
\end{table}

The gain over Reflexion comes primarily from optimization rather than additional correctness: the two methods have nearly identical correctness coverage (69.25\% vs.\ 69.34\%), while \methodname{}'s performance score increases from 25.90 to 31.77 and accounts for 79.2\% of the overall improvement (Table~\ref{tab:overall-score-decomposition}). Against OpenEvolve, runtime, correctness and performance all improve.

\obs{Observation~2.} Knowledge grounding serves \textbf{two roles}: it improves the quality of valid kernels and prevents search from being lost in invalid implementations.

\subsection{Does External Knowledge Help Beyond Experience?}
\label{sec:kb-ablation}

\begin{table}[t]
  \centering
  \begin{minipage}[t]{0.48\linewidth}
    \vspace{0pt}
    \caption{Level~3 knowledge-base ablation.}
    \label{tab:level3-kb-ablation}
    \centering
    \scriptsize
    \setlength{\tabcolsep}{2pt}
    \begin{tabular}{@{}p{0.40\linewidth}rrr@{}}
      \toprule
      \rowcolor{tableheadbg}
      Metric & No KB & With KB & $\Delta$ \\
      \midrule
      Build coverage               & 61.90\% & \textbf{100.00\%} & \posdelta{$\uparrow$\,38.10} \\
      Runtime-case coverage        & 57.86\% & \textbf{90.48\%}  & \posdelta{$\uparrow$\,32.62} \\
      Correct-case coverage        & 40.24\% & \textbf{59.76\%}  & \posdelta{$\uparrow$\,19.52} \\
      Performance-case coverage    & 17.86\% & \textbf{48.33\%}  & \posdelta{$\uparrow$\,30.48} \\
      Score $\uparrow$             & 29.49   & \textbf{56.94}    & \posdelta{$\uparrow$\,27.45} \\
      \bottomrule
    \end{tabular}
  \end{minipage}\hfill
  \begin{minipage}[t]{0.50\linewidth}
    \vspace{0pt}
    \caption{Stage-wise synthesis outcomes.}
    \label{tab:stage-wise-outcomes}
    \centering
    \scriptsize
    \setlength{\tabcolsep}{1.5pt}
    \begin{tabular}{@{}lrr@{}}
      \toprule
      Level & \thead[r]{Stage~2 vs. Stage~1\\relative gain (\%) $\uparrow$}
            & \thead[r]{Adoption\\rate (\%) $\uparrow$} \\
      \midrule
      L1 & 4.02 & 62.5 \\
      L2 & 3.26 & 87.5 \\
      L3 & 1.88 & 25.0 \\
      \bottomrule
    \end{tabular}
  \end{minipage}
\end{table}

Removing the knowledge base causes failures throughout the evaluation pipeline: build coverage falls from 100\% to 61.9\%, and performance coverage from 48.3\% to 17.9\% (Table~\ref{tab:level3-kb-ablation}). The effect is therefore not limited to making correct kernels faster. External knowledge allows substantially more implementations to survive compilation, execution and correctness validation before performance optimization can even begin. As a result, the L3 score nearly doubles from 29.49 to 56.94.

This effect is also visible in strategy diversity (Table~\ref{tab:correctness-effective-diversity}): the KB increases planned strategies only from 5.25 to 6.75 per operator, but increases full-correct strategies from \textbf{0.25} to \textbf{1.30}---a more than \textbf{5$\boldsymbol{\times}$} increase. Knowledge therefore does not merely make the agent propose more strategies; it makes a much larger fraction of distinct strategies survive platform and correctness constraints.

\obs{Observation~3.} Knowledge improves \textbf{\emph{effective}}, not merely nominal, exploration.

\subsection{When Does Mechanism-Level Synthesis Help?}
\label{sec:synthesis}

Stage~2 improves the final set over Stage~1 by 4.02\% on L1, 3.26\% on L2 and 1.88\% on L3 (Table~\ref{tab:stage-wise-outcomes}). Adoption is highest on L2 at 87.5\%, followed by L1 at 62.5\% and L3 at 25.0\%. The results indicate that synthesis is most effective when optimization mechanisms can be transferred while preserving validated semantics.

Stage~2 is effective when a donor mechanism matches a recurring input regime and can be integrated without changing the validated computation. Cases outside that regime retain the Stage~1 path and show little change. Crossover can be neutral or harmful when it addresses only a narrow correctness edge case, replaces a parallel implementation with a less efficient path, or introduces no substantive mechanism change. This pattern is consistent with the high adoption rate at L2 and the lower adoption and limited uplift at L3.
Appendix~\ref{app:stagewise-warmup-synthesis-cases} gives representative stage-wise case studies.

\obs{Observation~4.} Synthesis is most effective when mechanisms can be \textbf{composed without altering correctness-sensitive computation}.

\subsection{Does Knowledge Enable More Effective Exploration?}
\label{sec:diversity}

\begin{table}[t]
  \centering
  \begin{minipage}[t]{0.595\linewidth}
    \vspace{0pt}
    \caption{Archive-visible implementation diversity on 20 common Level~2--3 operators. \methodname{} is evaluated using its Stage~1 warmup candidates.}
    \label{tab:archive-visible-diversity}
    \centering
    \scriptsize
    \setlength{\tabcolsep}{2pt}
    \begin{tabular}{@{}lrrrr@{}}
      \toprule
      Method & \thead[r]{Strategy\\sig./op\\$\uparrow$}
             & \thead[r]{Kernel\\src./op\\$\uparrow$}
             & \thead[r]{Pairwise\\src. disagr.\\$\uparrow$}
             & \thead[r]{Pairwise\\struct. dist.\\$\uparrow$} \\
      \midrule
      \methodname{} (S1) & 6.75 & 13.90 & 95.1\% & 0.303 \\
      Reflexion          & 1.60 & 2.70  & 50.7\% & 0.081 \\
      OpenEvolve         & 1.30 & 1.65  & 19.6\% & 0.049 \\
      \bottomrule
    \end{tabular}
  \end{minipage}\hfill
  \begin{minipage}[t]{0.385\linewidth}
    \vspace{0pt}
    \caption{Correctness-effective planned-strategy diversity during warmup on 20 Level~3 operators.}
    \label{tab:correctness-effective-diversity}
    \centering
    \scriptsize
    \setlength{\tabcolsep}{2pt}
    \begin{tabular}{@{}p{0.555\linewidth}rr@{}}
      \toprule
      Metric & \thead[r]{With\\KB} & \thead[r]{No\\KB} \\
      \midrule
      Planned strategy sig. / op      & 6.75   & 5.25   \\
      Full-correct-eff. sig. / op     & 1.30   & 0.25   \\
      Full-correct strategy survival  & 19.3\% & 4.8\%  \\
      Ops with $\geq 1$ full-correct  & 40.0\% & 20.0\% \\
      \bottomrule
    \end{tabular}
  \end{minipage}
\end{table}

\methodname{} explores substantially more mechanism-distinct implementations than Reflexion and OpenEvolve (Table~\ref{tab:archive-visible-diversity}). More importantly, this diversity is not merely syntactic: with knowledge grounding, substantially more planned mechanisms survive correctness filtering (Table~\ref{tab:correctness-effective-diversity}). The KB increases full-correct strategy survival from 4.8\% to 19.3\%, supporting the role of the knowledge base as a source of viable mechanisms rather than simply additional prompting context.

\section{Related Work}
\label{sec:related}

\paragraph{LLM-based kernel generation.}
LLM-based kernel generation has progressed rapidly on CUDA and Triton.
\textsc{KernelBench}~\citep{ouyang2025kernelbench} established a systematic benchmark for functional correctness and performance, while subsequent work explored multi-agent refinement~\citep{zhang2025cudaforge,wei2025astra}, reinforcement learning with execution feedback~\citep{baronio2025kevin,li2025autotriton,woo2025tritonrl,zhu2025qimeng}, and evolutionary search~\citep{novikov2025alphaevolve,liao2025kernelevolve}.
Kernel generation is more challenging on specialized accelerator stacks such as Ascend~C, where general-purpose LLMs show limited out-of-the-box capability~\citep{cao2026ascendkernelgen}.
Recent work addresses this gap through domain adaptation~\citep{cao2026ascendkernelgen}, constrained intermediate representations~\citep{wen2025ascendcraft}, external knowledge augmentation~\citep{wen2026hawk,qiu2026agenticcann}, and reusable experience accumulated from previous search trajectories~\citep{zheng2026towards,wu2026ascendoptimizer}.
Our work focuses on the \textbf{complementarity} between the latter two directions.

\paragraph{Test-time search and experience for code agents.}
Test-time scaling improves agent performance through additional refinement, search, memory, or population-based exploration~\citep{madaan2023self,shinn2023reflexion,yao2023tree,wang2023voyager,novikov2025alphaevolve}.
For kernel generation, compilation and execution provide strong verification signals, motivating iterative and evolutionary search guided by previous outcomes~\citep{liao2025kernelevolve,zhang2026accelopt,zheng2026towards}.
However, increasing the number of trials does not necessarily increase the diversity of mechanisms explored when new candidates are generated primarily from previously discovered trajectories.
Our method \textbf{complements} experience-driven search with retrieved platform knowledge, using external evidence to seed mechanism-diverse implementation lines and execution feedback to evolve them.
Appendix~\ref{sec:more-related} relates our setting to automated kernel optimization before LLMs and to retrieval-augmented generation for code.

\section{Conclusion and Limitations}
\label{sec:conclusion}

\methodname{} combines trial experience with hierarchical, demand-driven platform knowledge for evolutionary kernel search. On 53 Ascend~C operators, it improves over Reflexion and OpenEvolve by 11.6\% and 17.2\%; ablations show that knowledge expands both the pool of candidates eligible for optimization and the diversity of correct mechanisms.

\paragraph{Limitations.}
Performance varies by difficulty: on the hardest operators, retrieved knowledge improves executability more than numerical correctness, but cannot guarantee semantic correctness.

\section*{AI Use Statement}

Large language models (LLMs) were used as general-purpose assistance tools in a limited capacity during the preparation of this work. Specifically, they assisted with grammar and spelling correction, implementation support for the proposed method and experiments, and manuscript formatting. The research ideas, methodological choices, experimental design, analysis and interpretation of results, and conclusions were developed and determined by the authors. All LLM-assisted code and textual revisions were independently reviewed, tested, and verified by the authors before being incorporated into this work. The authors take full responsibility for the scientific content and final presentation of the paper.

\bibliography{references}
\bibliographystyle{iclr2027_conference}

\clearpage
\appendix
\section{Knowledge Base Construction Details}
\label{app:kb}

Our knowledge base separates reusable platform knowledge from operator-specific context. The shared knowledge base $\mathcal{K}_G$ contains information reusable across operators, while the per-operator knowledge base $\mathcal{K}_o$ specifies the target task and identifies the most relevant evidence in $\mathcal{K}_G$.

\paragraph{Shared knowledge base.}
$\mathcal{K}_G$ contains 236 versioned sources drawn from four main categories (Table~\ref{tab:kb-sources}). These sources provide complementary information about platform APIs, implementation mechanisms, target-specific constraints and operator semantics. Each source is associated with its toolkit version and hardware scope. Version-mismatched sources can still provide useful implementation patterns, but are treated as leads rather than authoritative platform facts.

\begin{table}[t]
  \caption{Main sources of knowledge used by the search agent.}
  \label{tab:kb-sources}
  \centering
  \small
  \begin{tabular}{lp{6.4cm}}
    \toprule
    Knowledge source & Role \\
    \midrule
    Platform documentation and examples & API semantics, code patterns and build rules \\
    Target environment & Installed APIs, compiler behavior and versions \\
    Operator implementations & Tiling, vectorization, dtype and boundary handling \\
    Task and algorithm specifications & Workloads, semantics and correctness requirements \\
    \bottomrule
  \end{tabular}
\end{table}

\paragraph{Authority and version consistency.}
We separate semantic authority from platform authority. The executable reference determines what the operator must compute, while the installed toolkit and compiler determine which platform mechanisms are valid on the target system. Matching official sources are used next, while sources from other toolkit versions are treated as lower-priority implementation leads. This allows the agent to reuse useful implementation structure without inheriting incompatible APIs or task semantics.

\paragraph{Per-operator knowledge.}
Before search, an automatic preparation step constructs an operator-specific context $\mathcal{K}_o$. It contains a \emph{task contract}, describing the operator semantics, input regimes, target environment and evaluation requirements, and a \emph{retrieval brief}, which identifies relevant APIs, related implementations and known version or semantic conflicts in $\mathcal{K}_G$. The preparation step requires no manual annotation and does not modify candidate kernels. This two-level organization keeps platform knowledge reusable while exposing only a small, task-relevant subset.

\section{Retrieval Details}
\label{app:retrieval}

Retrieval is question-driven rather than operator-wide. At each search step, the agent formulates a concrete implementation question, such as whether an API supports a particular dtype or how a related operator handles a boundary condition. Retrieval then proceeds in three stages.

\paragraph{Route.}
The operator-specific retrieval brief maps the current question to a small set of relevant platform sources, related implementations and API entries.

\paragraph{Filter.}
Retrieved evidence is checked against the target toolkit and hardware. Version-mismatched sources may be retained as leads, but cannot override facts observed on the target.

\paragraph{Read.}
The agent reads the relevant source region directly, together with the local context required to interpret the implementation pattern. If needed, retrieval checks related sources.

The resulting evidence pack therefore contains a small set of source-grounded facts and implementation patterns relevant to the current decision, rather than a generic collection of documents that mention the operator.

\paragraph{Example.}
For \texttt{apply\_adam\_w}, retrieval identifies a closely related vendor implementation whose update rule differs from the executable reference in the placement of $\varepsilon$. The reference is treated as the authority for operator semantics, while the vendor implementation remains useful for its tiling and vectorization structure. The agent can therefore reuse the implementation mechanism without inheriting the semantic mismatch. This example illustrates why source identity and authority are preserved during retrieval: the agent borrows implementation structure from one source while adopting the arithmetic of another.

\section{More Related Work}
\label{sec:more-related}

\paragraph{Automated kernel optimization before LLMs.}
Generating fast kernels without a human expert is a long-standing goal.
Scheduling languages separate what a kernel computes from how it is executed, and search over the resulting schedule space, guided by auto-scheduling and learned cost models, has produced strong tensor programs~\citep{ragan2013halide,chen2018tvm,zheng2020ansor,zheng2021tenset,zhai2023tlp}.
Polyhedral compilers instead derive loop transformations analytically from a dependence model~\citep{bondhugula2008pluto,verdoolaege2013polyhedral,baghdadi2019tiramisu}, an approach that has also been carried to neural processing units~\citep{zhao2021akg}.
These systems search \emph{within} a space of transformations that the compiler fixes in advance, and the platform constraints that make a transformation legal are encoded in the compiler itself---an encoding that takes considerable expert effort to build for each new backend.
Our setting is the one where that encoding does not yet exist: the difficulty is less to tune a schedule that is already known to be valid than to find a valid implementation mechanism at all, and the space of candidate mechanisms is not enumerated in advance.
\methodname{} therefore recovers mechanisms and constraints from primary platform sources at search time, rather than from a transformation space defined beforehand.

\paragraph{Retrieval-augmented generation for code.}
Retrieval-augmented generation conditions a model on evidence fetched at inference time instead of relying on parametric memory alone~\citep{lewis2021retrieval}, and the idea transfers naturally to code: retrieving related snippets from a code base improves both generation and completion~\citep{parvez2021retrieval,lu2022reacc}.
Two refinements are especially relevant here.
Repository-level methods retrieve context scattered across files and iterate between retrieval and generation, using what has been written so far to drive the next query~\citep{zhang2023repocoder}.
Documentation-level methods retrieve API descriptions rather than code, which lets a model use libraries and functions it never saw during training~\citep{zhou2023docprompting}.
The second is the closer analogue to our setting: public code corpora are dominated by mainstream languages~\citep{kocetkov2022stack}, so a model's parametric knowledge of an emerging programming model is thin, while the corresponding documentation, headers and vendor implementations do exist and can be consulted.
Recent agents for Ascend accordingly inject external knowledge into generation~\citep{wen2026hawk,qiu2026agenticcann}.
Our use of retrieval departs from these formulations in three respects, each of which shapes the design in \S\ref{sec:retrieval}.
Queries are frequently exact symbols, so lexical identity matters more than embedding similarity; a passage drawn from a mismatched toolkit version is actively harmful rather than merely unhelpful, so evidence has to be filtered against the target environment before it is used; and retrieval is driven not by the partial program, as in repository-level completion, but by an explicit search-time question informed by failed trials.

\section{Mechanisms Across Diverse Implementations}
\label{app:diverse-implementation-mechanisms}

\begin{figure}[!ht]
  \centering
  \includegraphics[width=0.98\textwidth,trim=11pt 58pt 10pt 11pt,clip]{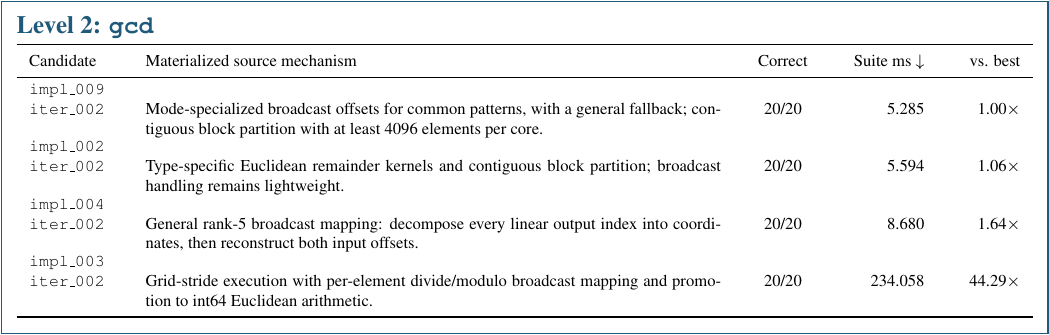}
  \caption{Diverse implementation mechanisms for \texttt{gcd}.}
  \label{fig:diverse-impl-gcd}
\end{figure}

\begin{figure}[!ht]
  \centering
  \includegraphics[width=0.98\textwidth,trim=11pt 62pt 10pt 11pt,clip]{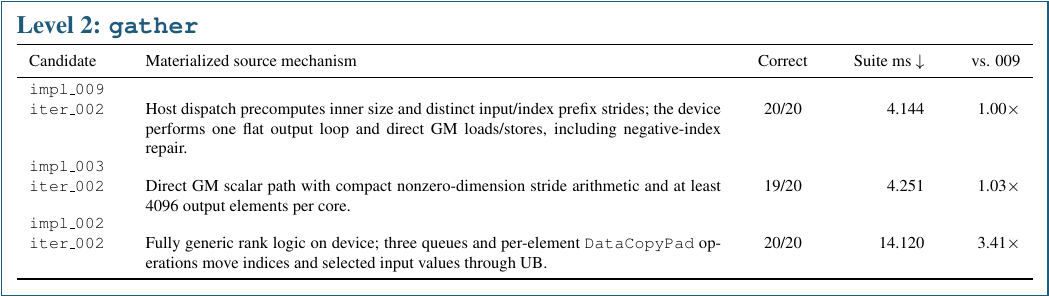}
  \caption{Diverse implementation mechanisms for \texttt{gather}.}
  \label{fig:diverse-impl-gather}
\end{figure}

\begin{figure}[!ht]
  \centering
  \includegraphics[width=0.98\textwidth,trim=11pt 34pt 10pt 11pt,clip]{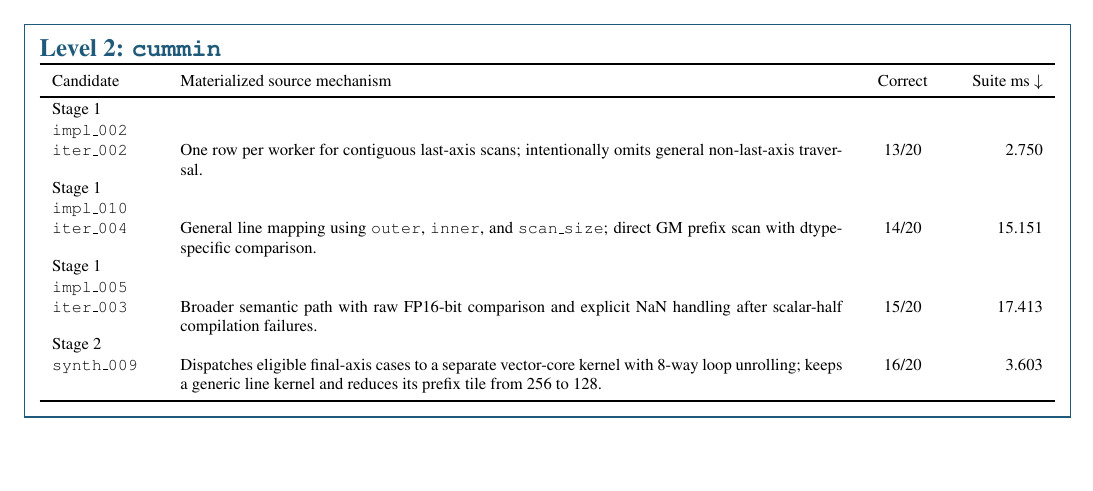}
  \caption{Diverse implementation mechanisms for \texttt{cummin}.}
  \label{fig:diverse-impl-cummin}
\end{figure}

\begin{figure}[!ht]
  \centering
  \includegraphics[width=0.98\textwidth,trim=11pt 58pt 10pt 11pt,clip]{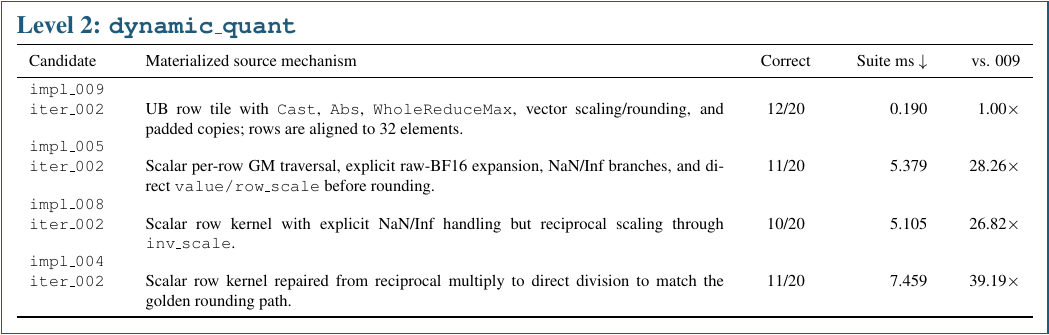}
  \caption{Diverse implementation mechanisms for \texttt{dynamic\_quant}.}
  \label{fig:diverse-impl-dynamic-quant}
\end{figure}

Figures~\ref{fig:diverse-impl-gcd}--\ref{fig:diverse-impl-dynamic-quant} show
the concrete execution mechanisms materialized by different archived
implementations. ``Correct'' is the number of benchmark cases passed. ``Suite
ms'' is the stage-local geometric mean over the candidate's passing cases,
using 5 warmups and 20 timed samples.

\section{AscendEvolve vs.\ Reflexion and OpenEvolve: Case Studies}
\label{app:correctness-case-studies}

\begin{figure}[!htbp]
  \centering
  \includegraphics[page=1,width=0.98\textwidth,trim=18pt 215pt 18pt 18pt,clip]{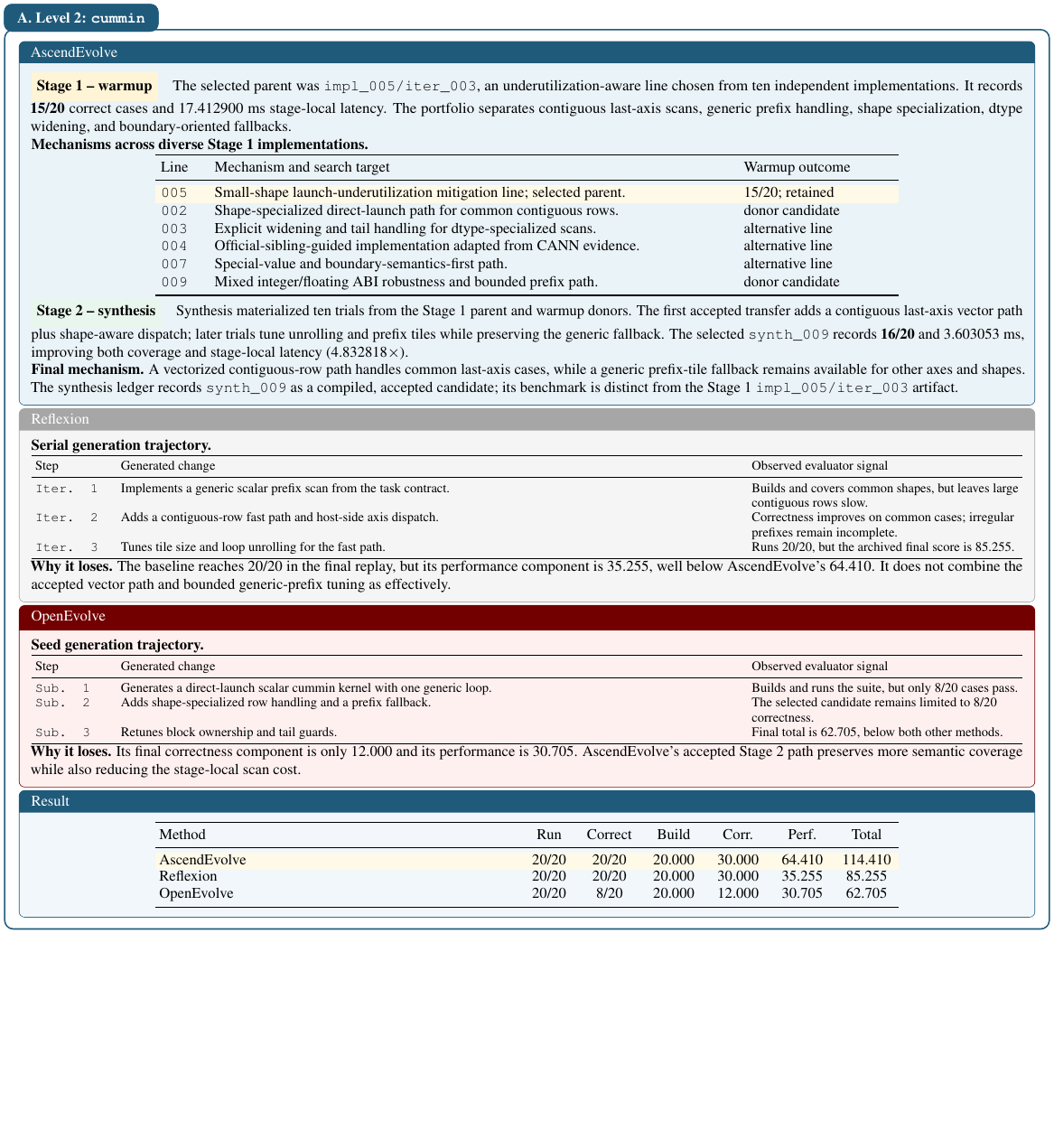}
  \caption{Case study for Level~2 \texttt{cummin}: Stage~2 combines a contiguous vector path with generic-prefix tuning.}
  \label{fig:case-cummin}
\end{figure}

\begin{figure}[!htbp]
  \centering
  \includegraphics[page=2,width=0.98\textwidth,trim=18pt 215pt 18pt 18pt,clip]{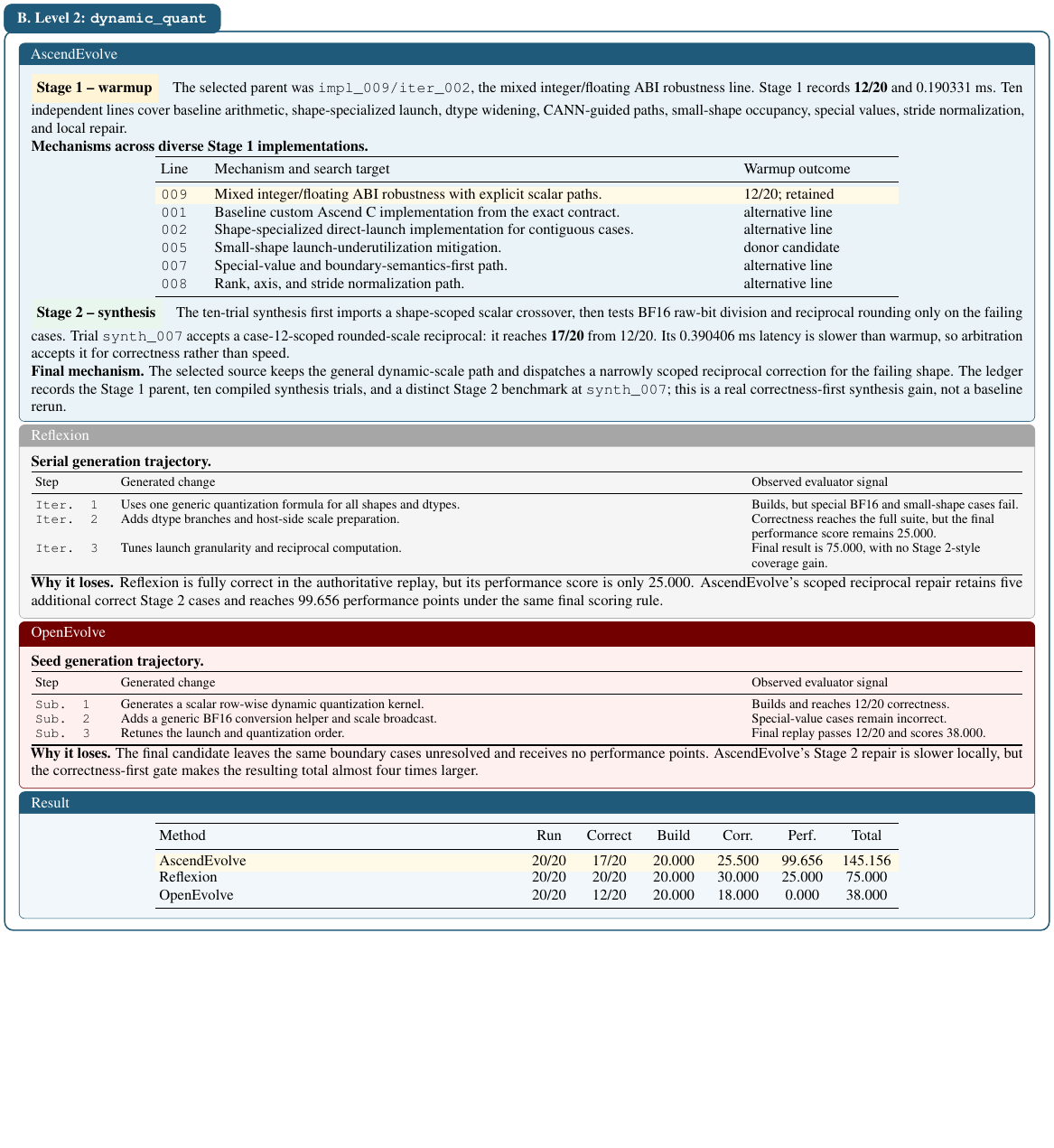}
  \caption{Case study for Level~2 \texttt{dynamic\_quant}: Stage~2 accepts a correctness-first scoped reciprocal repair.}
  \label{fig:case-dynamic-quant}
\end{figure}

\begin{figure}[!htbp]
  \centering
  \includegraphics[page=3,width=0.98\textwidth,trim=18pt 165pt 18pt 18pt,clip]{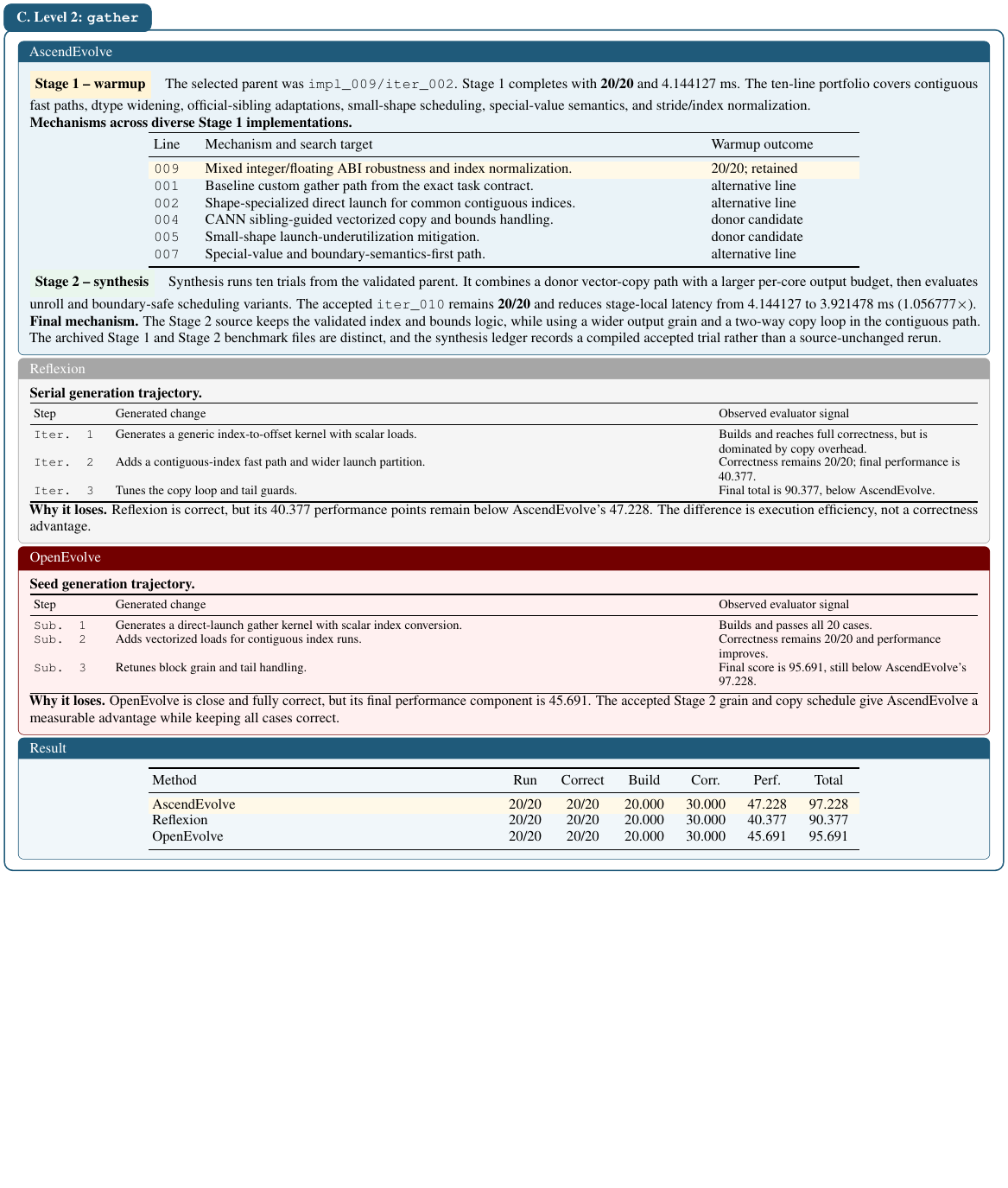}
  \caption{Case study for Level~2 \texttt{gather}: Stage~2 preserves 20/20 correctness while improving copy scheduling.}
  \label{fig:case-gather}
\end{figure}

Figures~\ref{fig:case-cummin}--\ref{fig:case-gather} compare the generation
trajectories and final scored outcomes of AscendEvolve, Reflexion, and OpenEvolve
on three Level~2 operators for which both Stage~1 warmup and Stage~2 synthesis
were actually materialized. Each case traces the mechanisms explored by
AscendEvolve and reports whether the advantage comes from performance,
correctness-first repair, or both.

\section{Stage-wise Case Studies of Warmup and Synthesis}
\label{app:stagewise-warmup-synthesis-cases}

\begin{figure}[!htbp]
  \centering
  \includegraphics[page=1,width=0.98\textwidth,trim=18pt 335pt 18pt 18pt,clip]{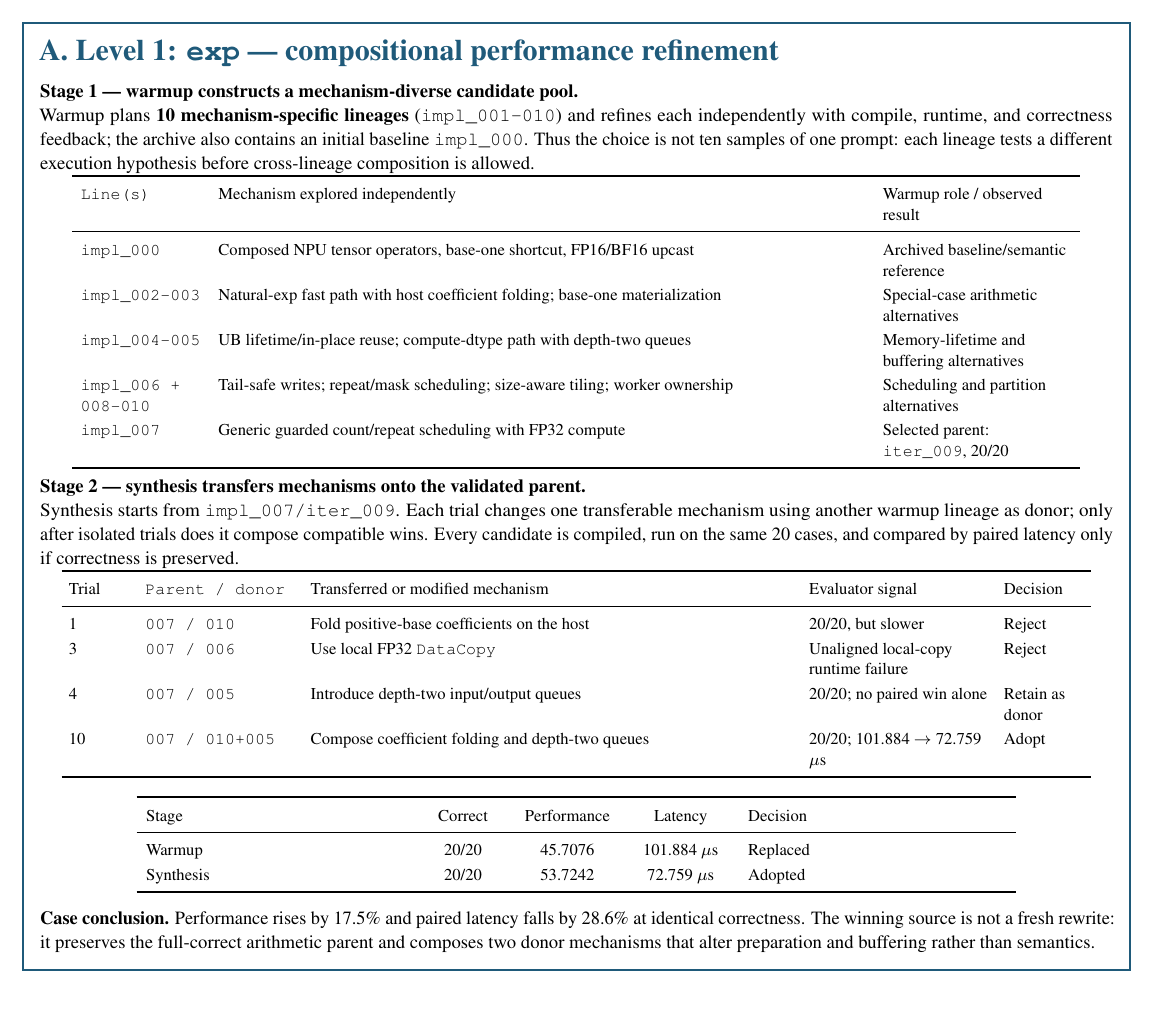}
  \caption{Stage-wise case study for Level~1 \texttt{exp}: synthesis composes coefficient folding and depth-two queues on a full-correct warmup parent.}
  \label{fig:stagewise-exp}
\end{figure}

\begin{figure}[!htbp]
  \centering
  \includegraphics[page=2,width=0.98\textwidth,trim=18pt 355pt 18pt 18pt,clip]{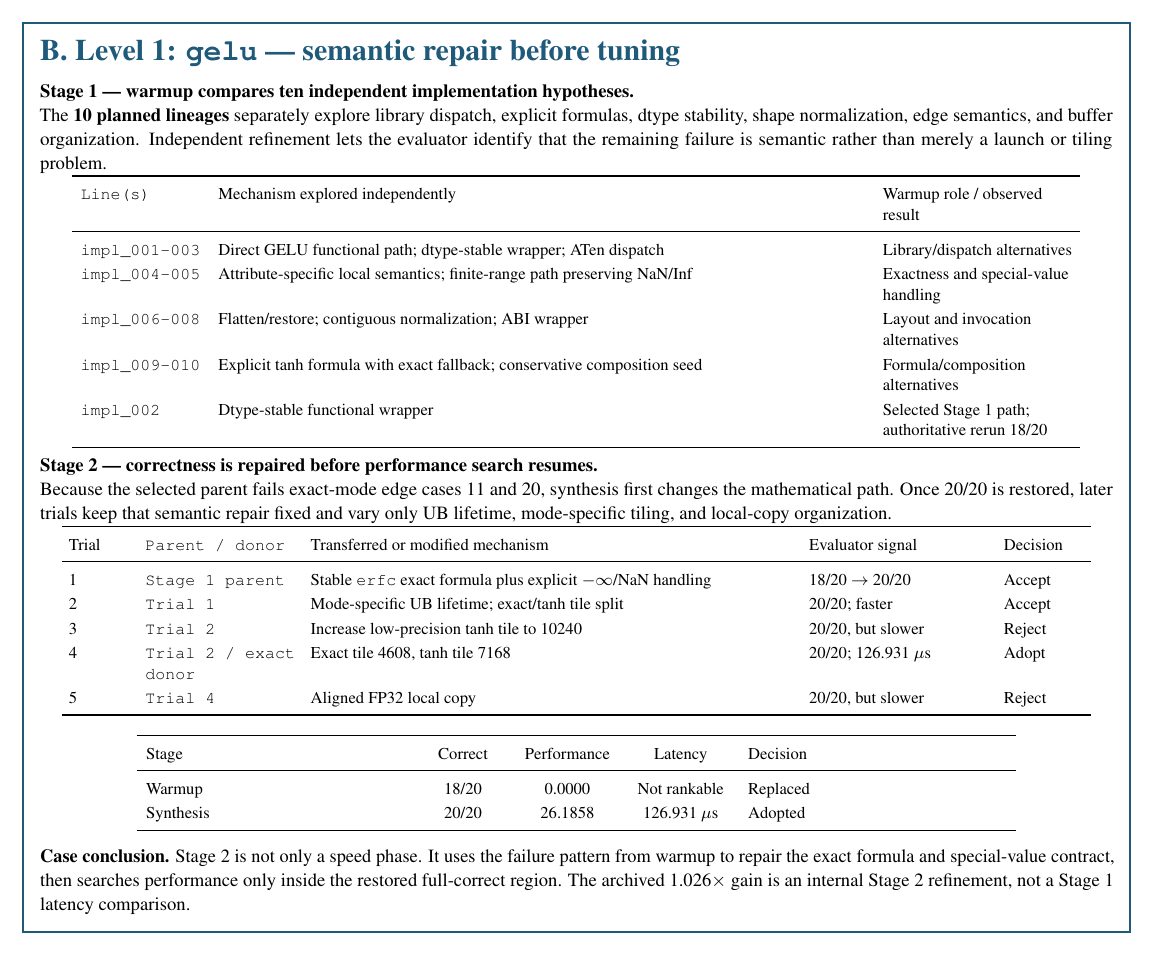}
  \caption{Stage-wise case study for Level~1 \texttt{gelu}: synthesis first repairs exact-mode semantics and then tunes within the full-correct region.}
  \label{fig:stagewise-gelu}
\end{figure}

\begin{figure}[!htbp]
  \centering
  \includegraphics[page=3,width=0.98\textwidth,trim=18pt 350pt 18pt 18pt,clip]{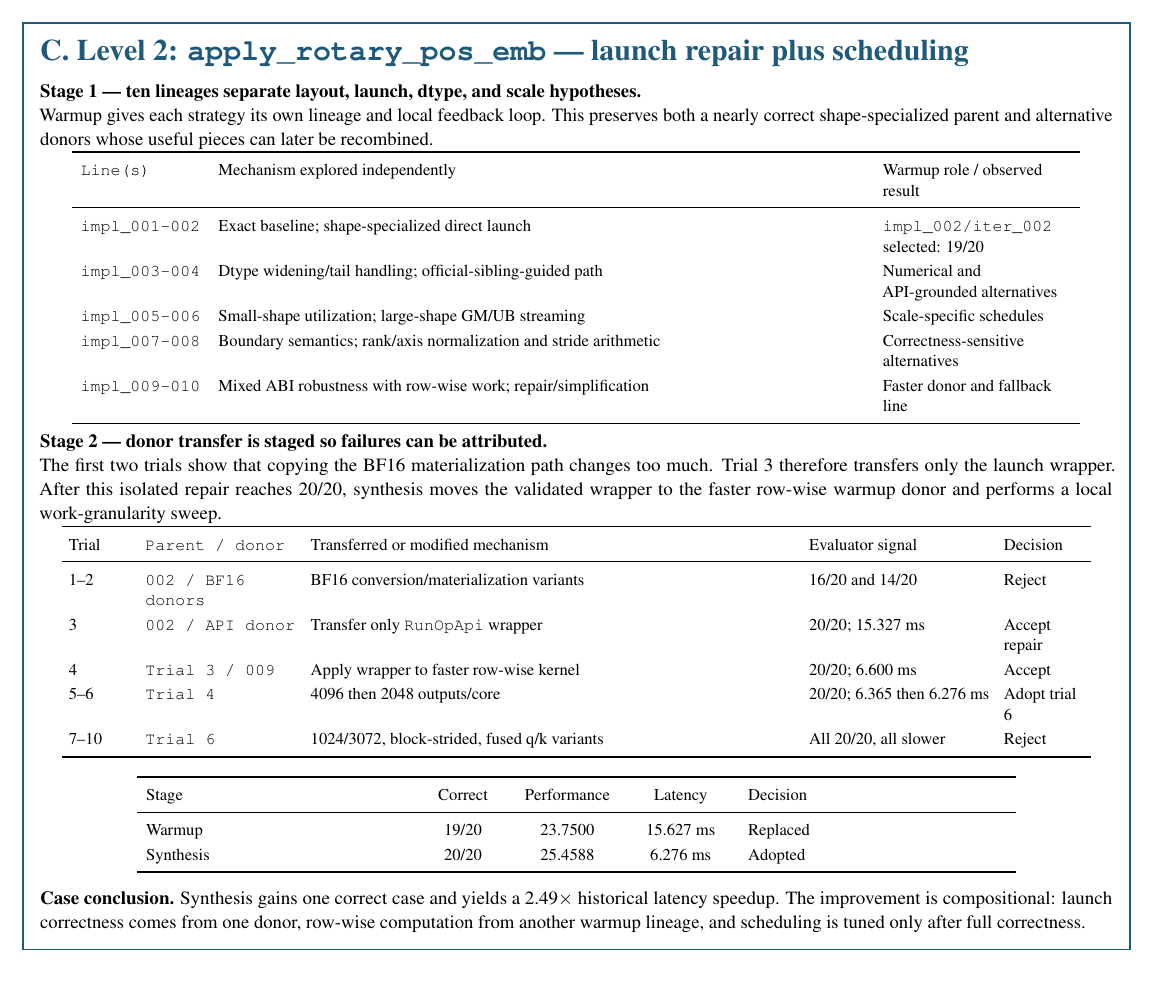}
  \caption{Stage-wise case study for Level~2 \texttt{apply\_rotary\_pos\_emb}: a launch-path repair is combined with a faster row-wise donor and scheduling sweep.}
  \label{fig:stagewise-rotary}
\end{figure}

\begin{figure}[!htbp]
  \centering
  \includegraphics[page=4,width=0.98\textwidth,trim=18pt 390pt 18pt 18pt,clip]{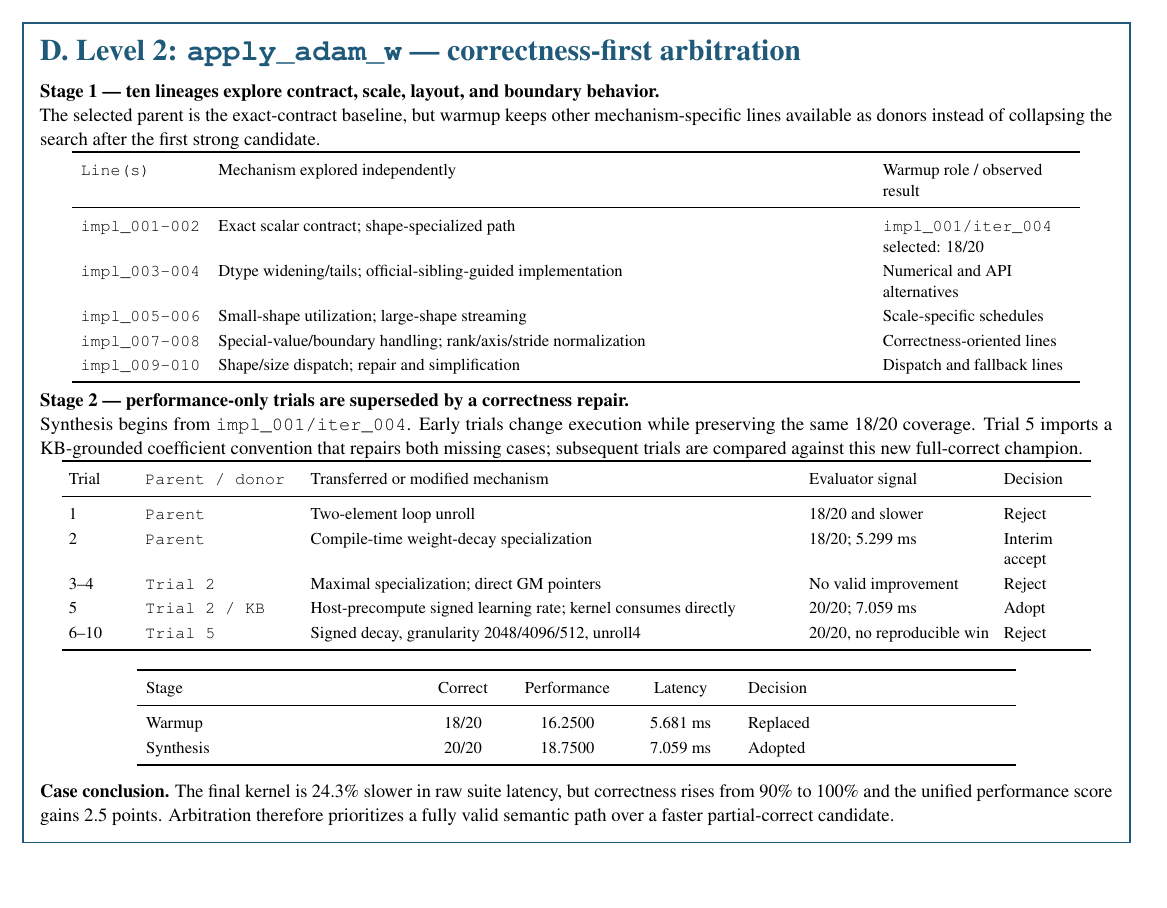}
  \caption{Stage-wise case study for Level~2 \texttt{apply\_adam\_w}: arbitration accepts a full-correct synthesis candidate despite higher raw latency.}
  \label{fig:stagewise-adamw}
\end{figure}

\begin{figure}[!htbp]
  \centering
  \includegraphics[page=5,width=0.98\textwidth,trim=18pt 385pt 18pt 18pt,clip]{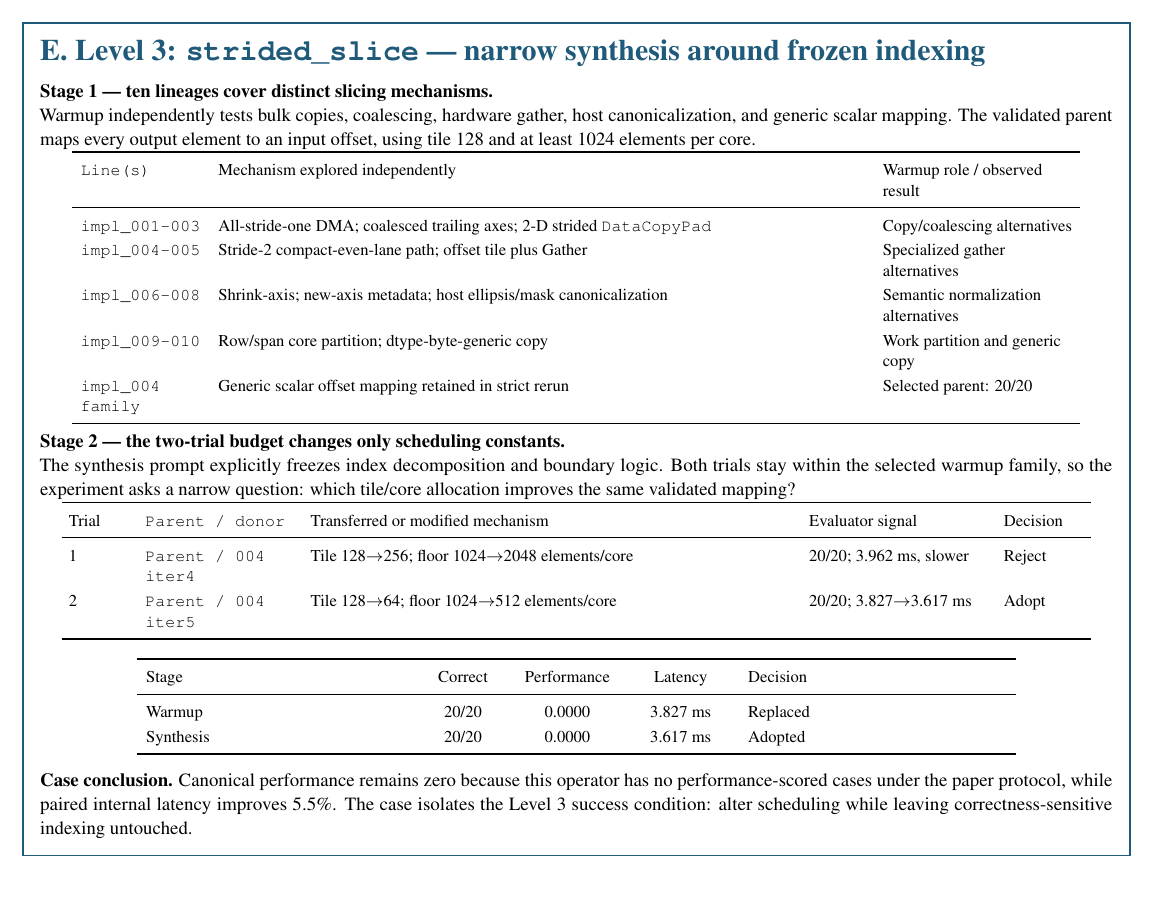}
  \caption{Stage-wise case study for Level~3 \texttt{strided\_slice}: synthesis changes scheduling constants while preserving validated indexing and boundary logic.}
  \label{fig:stagewise-strided-slice}
\end{figure}

\begin{figure}[!htbp]
  \centering
  \includegraphics[page=6,width=0.98\textwidth,trim=18pt 405pt 18pt 18pt,clip]{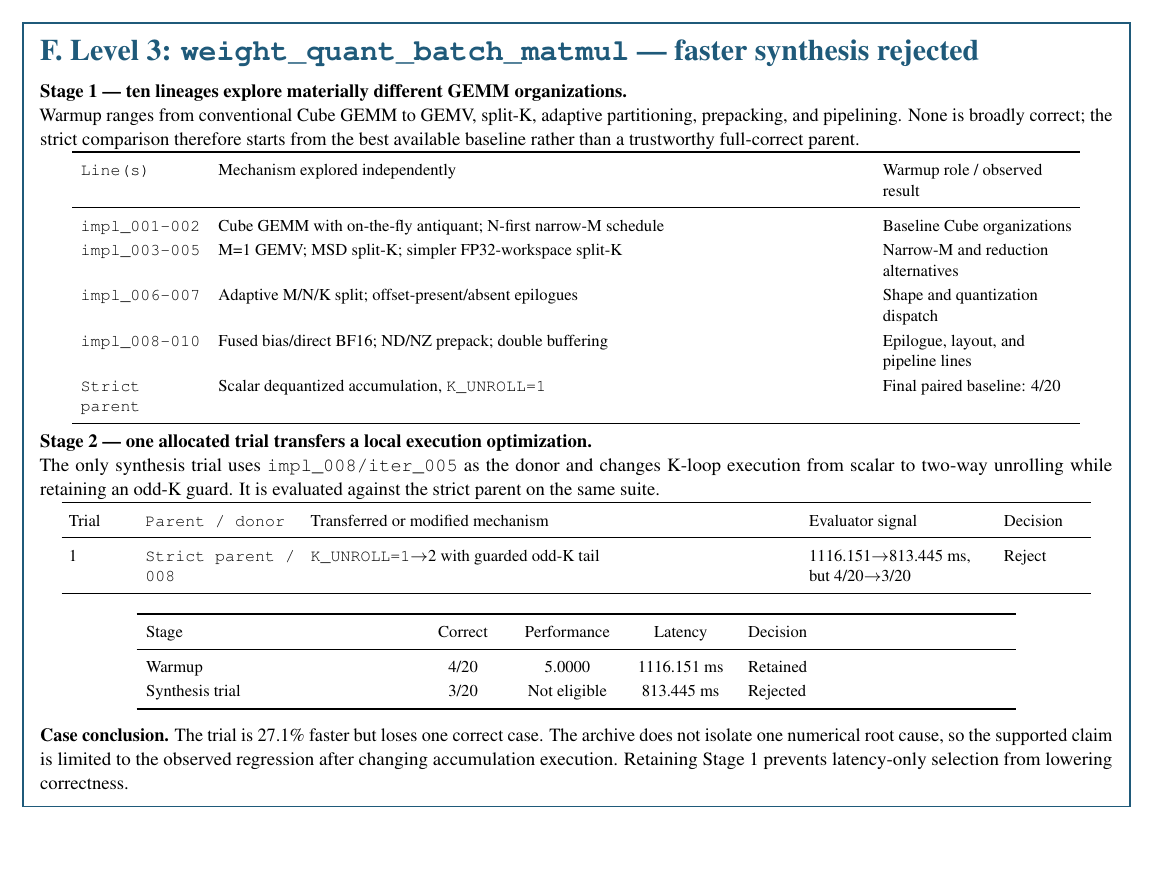}
  \caption{Stage-wise case study for Level~3 \texttt{weight\_quant\_batch\_matmul}: a faster synthesis trial is rejected after losing correctness coverage.}
  \label{fig:stagewise-weight-quant-bmm}
\end{figure}

\begin{figure}[!htbp]
  \centering
  \includegraphics[page=7,width=0.98\textwidth,trim=18pt 410pt 18pt 18pt,clip]{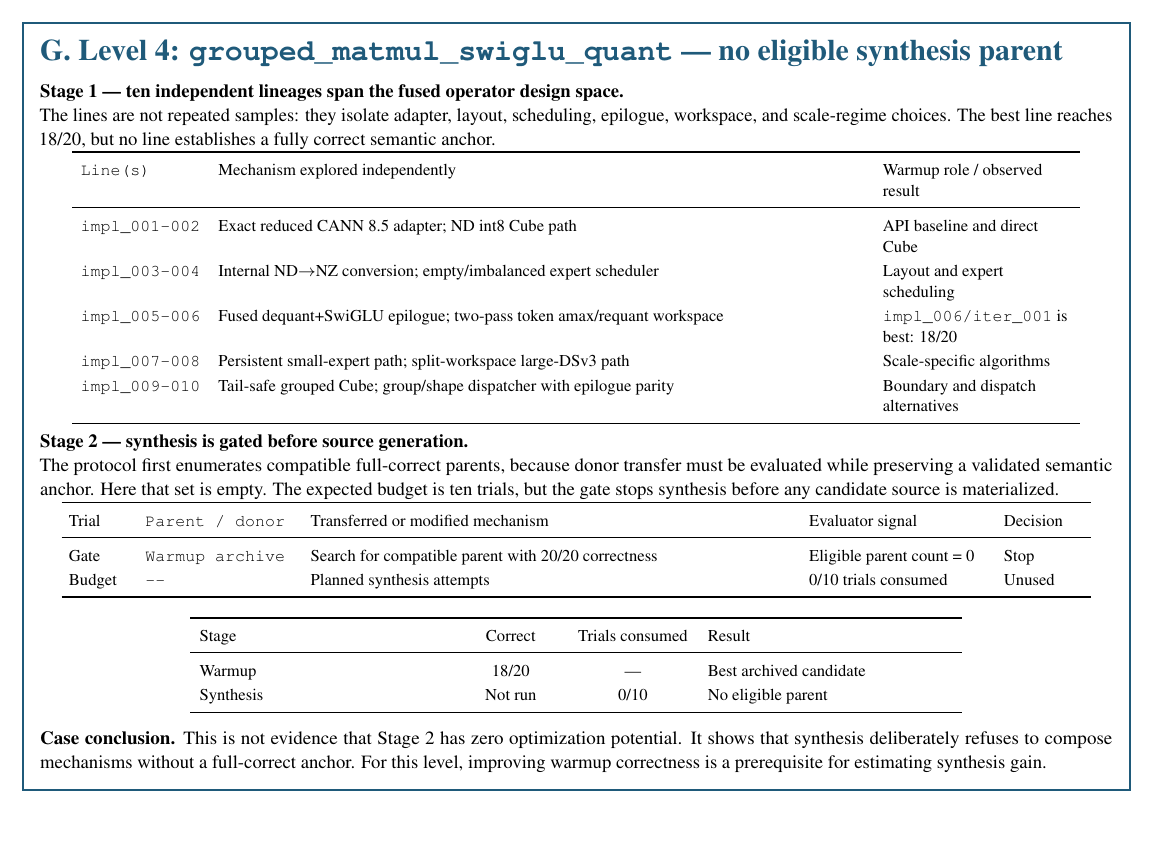}
  \caption{Stage-wise case study for Level~4 \texttt{grouped\_matmul\_swiglu\_quant}: synthesis is skipped because warmup provides no compatible full-correct parent.}
  \label{fig:stagewise-grouped-matmul}
\end{figure}

Figures~\ref{fig:stagewise-exp}--\ref{fig:stagewise-grouped-matmul}
trace how Stage~1 warmup constructs ten mechanism-specific implementation
lineages and how Stage~2 synthesis selects a warmup parent, transfers a donor
mechanism, and applies correctness-first arbitration. The cases cover
performance refinement, semantic repair, safe scheduling transfer, rejection
of a faster but less-correct candidate, and the absence of an eligible
full-correct synthesis parent.

\end{document}